\documentclass{article}                     
\usepackage{graphicx}
\usepackage{url}
\usepackage{algorithm2e}
\usepackage{amsmath}
\usepackage{amssymb}
\usepackage{latexsym}
\newtheorem{example}{Example}

\newcommand{\mytitle}{Preference Reasoning in Matching Procedures: Application to the Admission Post-Baccalaur\'eat Platform}

\begin{document}

\title{\mytitle
}


\author{Youssef Hamadi         \and
        Souhila Kaci 
}




\author{Youssef Hamadi$^{1}$ and Souhila Kaci$^{2}$\\
$^{1}$
Laboratoire d'Informatique (LIX) \\
\'Ecole Polytechnique
91128 Palaiseau Cedex, France\\
yhamadi75@gmail.com, youssefh@lix.polytechnique.fr  \\
$^{2}$LIRMM \\
University of Montpellier, Montpellier, France \\
kaci@lirmm.fr
}

\maketitle

\begin{abstract}
Because preferences naturally arise and play an important role in many real-life decisions, they are at the backbone of various fields. In particular preferences are increasingly used in almost all matching procedures-based applications. 
In this work we highlight the benefit of using AI insights on preferences in a large scale application, namely the French Admission Post-Baccalaur\'eat Platform (APB). Each year APB allocates hundreds of thousands first year applicants to universities. This is done automatically by matching applicants preferences to university seats. In practice, APB can be unable to distinguish between applicants which leads to the introduction of random selection. This has created frustration in the French public since randomness, even used as a last mean does not fare well with the republican egalitarian principle. In this work, we provide a solution to this problem. We take advantage of recent AI Preferences Theory results to show how to enhance APB in order to improve expressiveness of applicants preferences and reduce their exposure to random decisions.

\end{abstract}

\section{Introduction}


Preferences are everywhere in our daily lives. They occur as soon as we are faced
with a choice problem, e.g., ``which ice cream flavor would you prefer?'', ``which investment
funds would you choose?'', etc. Among multiple choices, it is often necessary
to identify one or more choices that are more appealing than others. Preference
is inherently a multidisciplinary topic which brings together artificial intelligence
researchers, philosophers, psychologists, economists, operations researchers, etc. In
particular, preferences are becoming of greater interest in many areas in artificial
intelligence, such as non-monotonic reasoning, multi-agent systems, constraint satisfaction,
decision making, social choice theory and decision-theoretic planning.

The last two decades have seen a widespread number of efforts dedicated to the problems of preferences in artificial intelligence ranging from their elicitation and representation to reasoning with/about preferences \cite{JuDeDoRoSc04,BiRo06,AI-Magazine08,Ka11,DoHuKaPr11}. More recently, we have witnessed an increasing interest in exploiting AI advances on preferences in other disciplines e.g. databases \cite{BrDo04,ToCi02,NeKa10} and game theory \cite{BoLaLaZa09}. There are still many domains where preferences techniques are integrated in a very primitive and naive way. This is the case in the French Admission Post-Baccalaur\'eat Platform (APB)\footnote{\url{https://www.admission-postbac.fr/}}. And in this work, our goal is to highlight the benefits of using AI insights in that large scale application.

The Baccalaur\'eat is an academic qualification which French students take at the end of high school. It was introduced by Napoleon Bonaparte in 1808. It is the main diploma required to pursue university studies. Since its inception, the number of laureates has dramatically increased, from 31 \textit{bacheliers} in 1809 to 632,700 in 2016, which have to choose between more than 12,000 university studies \cite{Piobetta,Figaro16}. These numbers, combined to the limited capacity of most universities make assignment procedures challenging. To make things worst, the vast majority of French universities have to respect a \emph{non-selection} principle which forbids the filtering of applicants based  on their previous high school results, Baccalaur\'eat options or motivation letters.

The adopted strategy exploits applicants expressed preferences to create institutions preferences. The idea is for an institution to prefer applicants who highly ranked it in their preferences list. This mechanism, along the collection of applicants preferences, and the execution of matching algorithms is implemented by the Admission Post-Baccalaur\'eat Platform. In practice, APB can be unable to distinguish between applicants. This is the case when they rank-order a study at the same level. APB's solution is to randomly order equivalent applicants in its preferences lists. Randomness, even used as a last mean does not fare well with the republican egalitarian principle, especially when it affects an institution like the \emph{bac}. Once officially acknowledged by the education minister, it has created frustration for the students and their parents, and as a consequence the education ministry has fixed an objective of reducing the use of randomness in APB \cite{APBStats}.

In this work, our aim is to tackle this problem and to provide concrete solutions to restrict the use of randomness in APB. We thoroughly analyse the APB platform and provide concrete improvements inspired by recent AI Preference Theory results. We show how and where APB can be fixed to implement our solutions and improve the performances of its allocation procedures. In particular, our work offers three benefits:

\begin{enumerate}
\item allow more flexibility for applicants to express their preferences (aiming for a better embedding of preferences), 
\item allow applicants to delay their definite preferences, 
\item reduce exposure to random decisions in matching algorithms.   
\end{enumerate}

The presentation is organized as follows. In Section 2, we give necessary background. Section 3 fully describes the APB algorithms and processes along general information on its practical use and perception. In Section 4, we show how to extend APB to improve preferences expressiveness, allow applicants to delay their definite preferences and reduce exposure to random decision making. Before giving a general conclusion in Section 6, Section 5 presents the related work.

\section{Preliminaries}\label{Sec:prem}
\subsection{Preferences Modeling}
Preferences modeling is based on a finite set of objects, denoted by $\mathcal{O}$, to be compared or evaluated \cite{Fi70}. The basic ingredient in this framework is a binary relation, denoted by $\succeq$, over $\mathcal{O}$. The notation $o\succeq o'$ stands for ``$o$ is at least as preferred as $o'$''. Thus $\succeq$ is referred to as {\it preference relation}.\\

Given a preference relation $\succeq$ and two objects $o,o'\in\mathcal{O}$, we distinguish between three relations over $o$ and $o'$:
\begin{itemize}

\item $o$ is strictly preferred to $o'$, denoted by $o\succ o'$, when $o\succeq o'$ holds but $o'\succeq o$ does not. $\succ$ is called a {\it strict preference relation}. 

\item $o$ is indifferent to $o'$, denoted by $o\approx o'$, when both $o\succeq o'$ and $o'\succeq o$ hold.\linebreak 
$\approx$ is called an {\it indifference relation}. 

\item $o$ is incomparable to $o'$, denoted by $o\sim o'$, when neither $o\succeq o'$ nor $o'\succeq o$ holds.\linebreak 
$\sim$ is called an {\it incomparability relation}.

\end{itemize}

Let us now recall some basic properties concerning binary relations.
\begin{itemize}

\item $\succeq$ is reflexive if and only if $\forall o\in\mathcal{O}$, $o\succeq o$.

\item $\succeq$ is irreflexive if and only if $\forall o\in\mathcal{O}$, $o\succeq o$ does not hold.

\item $\succeq$ is complete if and only if $\forall o,o'\in\mathcal{O}$, we have $o\succeq o'$ or $o'\succeq o$.

\item $\succeq$ is transitive if and only if $\forall o,o',o''\in\mathcal{O}$, if $o\succeq o'$ and $o'\succeq o''$ then $o\succeq o''$.

\item $\succeq$ is symmetric if and only if $\forall o,o'\in\mathcal{O}$, if $o\succeq o'$ then $o'\succeq o$.

\item $\succeq$ is antisymmetric if and only if $\forall o\in\mathcal{O}$, $\forall o'\in\mathcal{O}\backslash\{o\}$, we have  $not(o\succeq o'\mbox{ and }o'\succeq o)$.

\item $\succeq$ is asymmetric if and only if $\forall o,o'\in\mathcal{O}$, we have $not(o\succeq o'\mbox{ and }o'\succeq o)$. 

\end{itemize}

Given the properties of a preference relation $\succeq$, we distinguish between different types of preference structures. For the work presented here, we need to recall two structures:
\begin{itemize}

\item {\it A total preorder:} this corresponds to a reflexive, complete and transitive preference relation $\succeq$. The associated incomparability relation is empty.\\
When $\succeq$ is antisymmetric, $\approx$ is the set of pairs $(o,o)$ and the preference structure is called a {\it total order}. Lastly, when $\succeq$ is asymmetric, $\approx$ is empty and the preference structure is called a {\it strict total order}.  

\item {\it A partial preorder:} this corresponds to a reflexive and transitive preference relation $\succeq$. The associated incomparability relation is not empty.\\
When $\succeq$ is antisymmetric, $\approx$ is composed of pairs $(o,o)$ only and the preference structure is called a {\it partial order}. Lastly, when $\succeq$ is asymmetric, $\approx$ is empty and the preference structure is called a {\it strict partial order}. 

\end{itemize}
A preference relation $\succeq$ is cyclic if and only if its induced strict preference relation $\succ$ is cyclic, i.e., there exists a chain of objects $o,\cdots,o'$ such that $o\succ\cdots\succ o'\succ o$. Otherwise $\succeq$ is acyclic.\\

From now on, we suppose that $\succeq$ is transitive. By abuse of language we sometimes say that $\succeq$ is a total or partial (pre)order or (strict) total or partial (pre)order. When no confusion is possible, an order is equivalently  denoted by $\succeq$ or $\succ$.\\

When the preference relation $\succeq$ is a total preorder, the indifference relation $\approx$ induced by $\succeq$ is an equivalence relation (reflexive, symmetric and transitive). The set of equivalence classes of $\mathcal{O}$ given $\approx$ is totally ordered w.r.t. $\succ$. Let $E_1,\cdots,E_n$ be the set of equivalence classes induced by $\approx$. Then, 
\begin{description}

\item (i) $\forall i=1,\cdots,n$, $E_i\neq\emptyset$,
 
\item (ii) $E_{1}\cup\cdots\cup E_{n}=\mathcal{O}$, 

\item (iii) $\forall i,j$, $E_{i}\cap E_{j}=\emptyset$ for $i\neq j$, 

\item (iv) $\forall o,o'\in E_i$, $o\approx o'$.

\end{description}
$(E_1,\cdots,E_n)$ is an ordered partition of $\mathcal{O}$ given $\succeq$ iff ($\forall o,o'\in\mathcal{O}$, $o\in E_i$, $o'\in E_j$ with $i<j$ if and only if $o\succ o'$). 
\begin{example}\label{ex}
Let $\mathcal{O}=\{o_0,o_1, o_2,o_3\}$ be a set of objects. Let $\succeq$ be 
a total preorder over $\mathcal{O}$ defined by $o_1\succ o_3$, $o_3\approx o_0$ and
$o_3\succ o_2$. Then, the ordered partition of $\mathcal{O}$ is
$(E_{1},E_{2},E_{3})$ with $E_{1}=\{o_1\}$, $E_{2}=\{o_0,o_3\}$ and $E_{3}=\{o_2\}$.\\
\end{example}
A total order can also be written as an ordered partition of $\mathcal{O}$ where each equivalence class is composed of a single object.

Note that an ordered partition of $\mathcal{O}$ associated with $\succeq$ is acyclic. For example, the preference relation $o_1\succ o_2$, $o_2\succ o_3$, $o_3\succ o_2$ and $o_3\succ o_4$ cannot be written in terms of an ordered partition.\\

Let $\succeq$ and $\succeq'$ be two preference relations. 
We say that $\succeq$ extends $\succeq'$ if and only if $\forall o,o'\in\mathcal{O}$, if $o\succeq' o'$ 
then $o\succeq o'$.
\begin{example}(Example \ref{ex} continued)\\
Let $\succeq':o_1\succ o_0$, $o_1\succ o_3\succ o_2$. Then $\succeq: o_1\succ o_3$, $o_3\approx o_0$, $o_3\succ o_2$ extends $\succeq'$. 
\end{example}

\subsection{Matching Algorithms}
The Stable Marriage Problem (SMP) aims at finding a matching between $2n$ women and men, using preferences lists (strict total orders) in which each person has expressed her preference over the members of the opposite gender. The matching must be stable which means that there is no married pair that would be individually better off than they are with the element to which they are currently matched. 

This problem was first described by D. Gale and L. Shapley as a special case of the more general College Admission problem \cite{gale62a}. In this seminal work, it is shown that it is always possible to solve the SMP and make all marriages stable using an $O(n^2)$ procedure presented in Algorithm \ref{Alg:SM}.

\begin{algorithm}[htbp]
\KwData{M list of $n$ men, W list of $n$ women}
\KwResult{$P$ set of $n$ engagements}
\Begin{
  Initialize all $m \in M$ and $w \in W$ to free\\
  Initialize $P$ to $\varnothing$\\
  \While {$\exists$ a free man $m$ who still has a woman $w$ to propose to}
  {
  \eIf {$w$ is free}
  {// $m$ and $w$ become engaged\\
  add $(m,w)$ to $P$}
  {// some pair $(m', w)$ already exists\\
  \eIf {$w$ prefers $m$ to $m'$}
  {
  // $m$ and $w$ become engaged\\
  remove $(m',w)$ from $P$ \\
  $m'$ becomes free\\
  add $(m,w)$ to $P$
  }
  {// $m'$ and $w$ remain engaged}
  }
}
\KwRet{$P$}
\label{Alg:SM}
}
\caption{Gale and Shapley Stable Marriage}
\end{algorithm}

The algorithm runs a sequence of proposals, and at each point in the process, each person is either \emph{engaged} or \emph{free}. Assuming proposals from men to woman, each man may alternate between being engaged and being free. If a man is engaged more than one time, his successive partners are less and less desirable. Once a woman is engaged, she will never be free again. She remains engaged, possibly with different partners.

In each sequence, a free man $m$ can propose the woman $w$ at the top of his preference list. If $w$ is free, she becomes engaged. If her current partner $m'$ is such that $m' \succ m$ in her preference list, she remains engaged with  him. Otherwise, she engages with $m$, and $m'$ is freed. The process stops when all men are engaged and returns a list of $n$ engagements. The solution is stable otherwise there are a non paired man $m$ and woman $w$ who prefer each other over their partners. If this was the case, $m$ would have proposed $w$ before his current partner, which gives a chance to $w$ to have engaged $m$ before rejecting him, and therefore there is a man $m'$ that she prefers to him.

Remark that the algorithm is optimal for those who propose but not necessarily for those who choose, i.e.,  Algorithm \ref{Alg:SM} respects men preferences best.

\begin{example}[Borrowed from \cite{gale62a}]
Let $\alpha$, $\beta$ and $\gamma$ be three men, and $A$, $B$ and $C$ be three women. Table \ref{tab} gives men and women preferences. 
\begin{table}
\center
\begin{tabular}{  c || c | c | c  }
 & $A$ & $B$ & $C$ \\ \hline \hline
$\alpha$ &   (1,3) & (2,2) & (3,1) \\
$\beta$ &    (3,1) & (1,3) & (2,2) \\
$\gamma$ &  (2,2) & (3,1) & (1,3) \\
 \end{tabular}\caption{Example of Stable Marriage}\label{tab}
\end{table}
For each pair $(a,b)$ in the table $a$ refers to the ranking of the woman by the man and $b$ given the ranking of the man by the women. For example $\alpha$ prefers $A$ first, $B$ second and $C$ third, while $B$ prefers $\gamma$ first, $\alpha$ second and $\beta$ third.\\
Three stable marriages are possible: $P_1=\{(\gamma,A),(\alpha,B),(\beta,C)\}$, \linebreak
$P_2=\{(\alpha,A),(\beta,B),(\gamma,C)\}$ and $P_3=\{(\beta,A),(\gamma,B),(\alpha,C)\}$. In $P_1$ both men and women have their second choice. In $P_2$ men have their first choice while in $P_3$ women have their first choice. Algorithm \ref{Alg:SM} returns $P_2$ as it gives priority to men.  
\end{example}


\section{The Admission Post-Baccalaur\'eat Platform}
\subsection{An Interactive Process}\label{subsec-description-APB}

The post-Baccalaur\'eat platform manages the interaction between applicants and institutions. It allows the collect of their preferences, and executes different allocation algorithms in an interactive multi stages process which run between January and July each year. The algorithms consider three classes of studies: selective, non selective with limited capacity, and unlimited capacity. The APB process is described in Figure \ref{Fig:APB}. It follows three main steps: 

\begin{figure}[htbp]
\begin{center}
   \includegraphics[scale=.7]{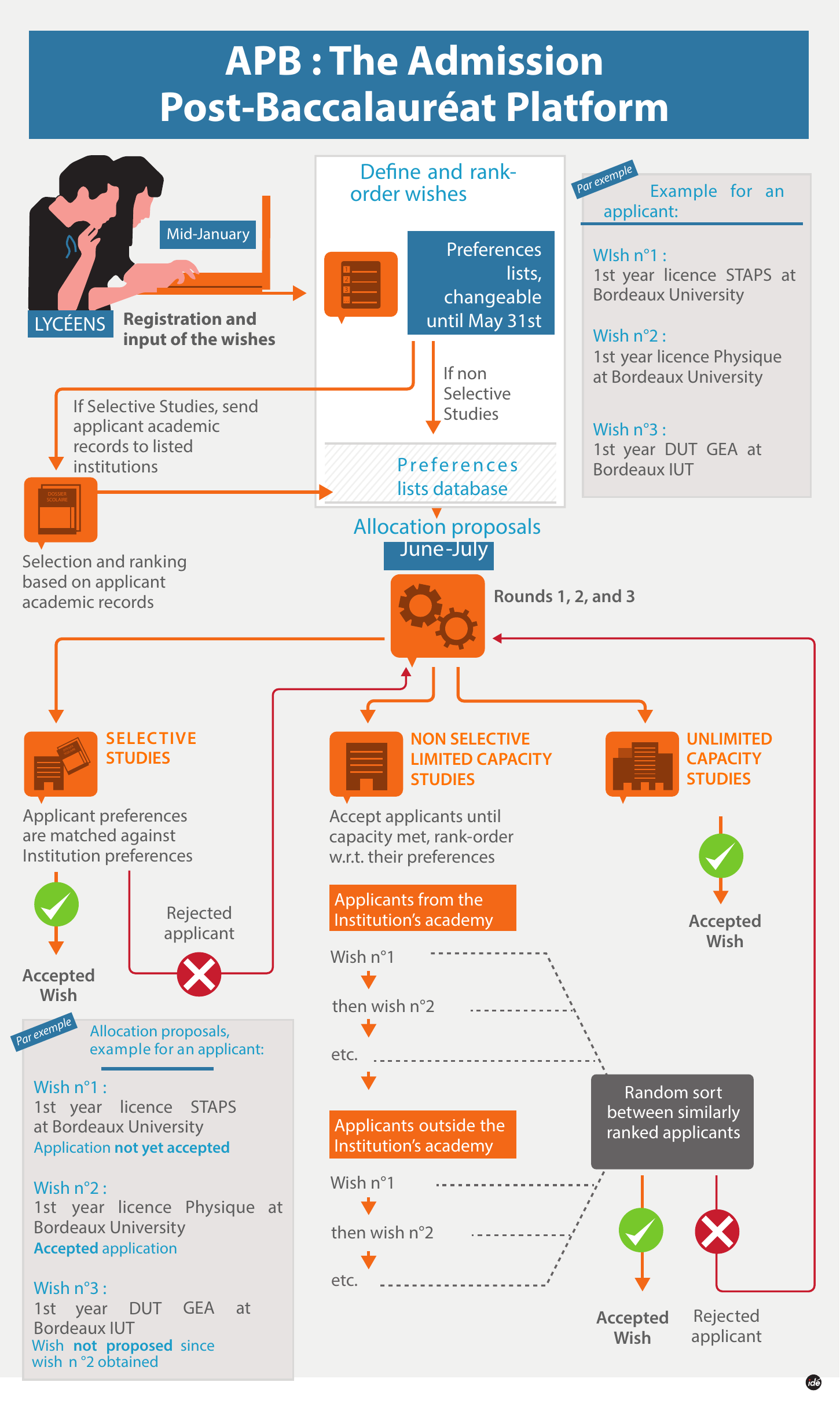}{\centering}
   \caption{\label{Fig:APB} The Admission Post-Baccalaur\'eat Platform (APB). (source: Ministry of National Education, France)}
\end{center}
\end{figure}

\begin{description}

\item {\bf Step 1: [Applicants' preferences]} 
Applicants express their preferences by means of a ranked list (strict total order) of wishes of studies, called preferences lists. Then the APB platform sends applicants academic records to selective studies (if any) for which they would have applied. 

\item {\bf Step 2: [Institutions' preferences]} Three classes of studies are considered:
\begin{itemize}

\item[] {\it Selective studies:} An institution with a selective study establishes a rank-ordered list (strict total order) of applicants according to its own criteria and based on applicants academic records. 

\item[] {\it Studies with limited capacity:} These institutions create their preferences by using applicants preferences. They give priority to applicants preferring them first. This is the only legal way to distinguish between applicants in order to respect limited capacity constraints. The top-k applicants are accepted until capacity is reached. The algorithm used to create institutions' preferences is described in Section \ref{sec:algo:rank}. 

\item[] {\it Studies with unlimited capacity:} All applications are accepted.  

\end{itemize} 

For each wish expressed in an applicant's preferences list given in Step 1, we have four possible answers: (1) Accepted wish, (2) Application not yet accepted, (3) Wish not proposed because of a better ranked proposed wish, and (4) Rejected applicant.

\item {\bf Step 3: [Allocation proposals]} Students are asked to accept (or not) proposed assignments (called proposals). APB makes proposals to applicants in three rounds. Only one proposal is offered to applicants in each round. APB takes into account:
\begin{itemize}
\item Applicants preferences lists.
\item Preferences lists established by institutions with selective studies or studies with limited capacity. 
\item and, from round 2, positions released by other applicants in previous rounds. 
\end{itemize}
 
Four possible answers are offered to applicants:
\begin{itemize}
\item {\it Definitely Yes:} The applicant definitely accepts the proposal. Therefore,
\begin{itemize}
\item no other proposal will be made for the applicant in subsequent rounds, 
\item if the applicant gives up this accepted proposal later in the process then she has to resign.
\end{itemize}
\item {\it Yes but:} The applicant accepts for the moment the proposal with the hope of getting a better proposal in subsequent rounds (i.e. she will be possibly offered a better ranked wish in subsequent rounds). Therefore, if in a subsequent round the applicant is offered a better ranked wish then the previous proposal is definitely deleted.
\item {\it No but:} The applicant refuses the proposal but maintains her applications for better ranked wishes. Therefore, the refused proposal is definitely deleted.

\item {\it Resign:} The applicant gives up (renounces) all her applications. 
\end{itemize}

\end{description}


\subsection{Use of Matching Algorithms in the APB platform}\label{sec:algo:rank}

The Stable Marriage procedure presented in Section \ref{Sec:prem} is used to match applicants to institutions for both Selective and Non Selective Limited Capacity Studies. The APB's implementation of this procedure is institution-proposing, i.e., it respects institutions preferences best\footnote{Julien Grenet, personal communication.}.

Selective Studies can enforce their own criterion to rank-order applicants. The SM procedure is then fed with the preferences of the institutions and preferences of the applicants.

On the other hand, institutions running Non Selective Limited Capacity Studies are not legally allowed to filter applicants based on their previous high school results, Baccalaur\'eat options or motivation letters. Still, their limited capacity requires some selection process. This is done by using applicants preferences to create the preferences lists of institutions. 
More concretely, it starts by first giving priority to applicants of the study academy. These are applicants who take their Baccalaur\'eat in the academy and/or live in the academy. These applicants are then rank-ordered according to their preferences expressed in Step 1. The higher ranked a study in an applicant's preferences list, the better the applicant rank is for this particular study. Then applicants outside of the academy are considered. Before we formally describe the APB algorithm, we present two notions used in the algorithm namely {\it absolute rank wish} and {\it relative rank wish}. An absolute wish refers to the rank of the study in the applicant preferences list. As studies are grouped into families, a relative wish corresponds to the rank of a given study w.r.t. studies pertaining to the same family. An example is given below.

\begin{example}
Suppose we have the following preferences list of a given applicant:
\begin{description}
\item Wish 1: Ecole Pr\'epa Lyc\'ee 1 (CPGE)
\item Wish 2: Licence Physique (LMD)
\item Wish 3: Ecole Pr\'epa Lyc\'ee 2 (CPGE)
\item Wish 4: Licence Maths (LMD)
\end{description}

Licence Maths is fourth in absolute ranking and second in relative (LMD) ranking.
\end{example}

We can remark that the distinction between absolute and relative ranking is a naive way to approximate incomparability between groups of objects. See Section \ref{Sec:prem}.

\begin{algorithm}[htbp]
\KwData{\emph{Institutions}, list of institutions}
\KwResult{For each study $s$ belonging to some institution $i$, \emph{s.Selected} preferences list of applicants}
\Begin{   
  \For {i $\in$ Institutions}{  
    \For {s $\in$ i.Studies}{ 
    		\For(\tcp*[h]{relative ordering}){r $\in$ 1 .. 24}{  
    			\For(\tcp*[h]{absolute ordering}) {a $\in$ 1 .. 24}{  
				locals = $\{ app \in s.Applicants$ s.t. \\
					$app.Academy = i.Academy$
					and\\
					$app.relative(s) = r$
					and
					$app.absolute(s) = a \}$\\
				$randList$ = randomSort(locals)\\
				$add(s.Selected, randList)$ 
			}
  		}
  		
   		\For(\tcp*[h]{relative ordering}){r $\in$ 1 .. 24}{  
    			\For(\tcp*[h]{absolute ordering}) {a $\in$ 1 .. 24}{
    				nonLocals = $\{ app \in s.Applicants$ s.t. \\
					$app.Academy \neq i.Academy$
					and\\
					$app.relative(s) = r$
					and
					$app.absolute(s) = a \}$\\
				$randList$ = randomSort(nonLocals)\\
				$add(s.Selected, randList)$
			}
  		}
  }
  }
\label{Alg:CreatePrefOne}
\caption{Create Institutions Preferences}
}
\end{algorithm}

Algorithm \ref{Alg:CreatePrefOne} presents this processing. It is based on the description given by the French ministry of education \cite{APBAlgoCrit}. It exploits absolute and relative ranks to separate applicants and create for each study a totally ordered preferences list.

The method iterates over institutions and their studies, successively considering local and non local applicants. It creates sets \emph{locals} (resp. \emph{nonLocals}) of equivalents applicants, i.e., members of a set have ranked the considered study at the same relative and absolute ranks. Each set is transposed into a \emph{randList} list data structure which randomly orders applicants. This local list is then appended to the study's \emph{Selected} list. This structure gives for each study a totally ordered list of selected applicants.\\

\subsection{Practical Usage and Perception}


In 2015, 788,000 applicants used the APB platform\footnote{A 5\% increase against the previous year.} to select among more than 12,000 studies. Among them, 38\% put a non selective limited capacity study as their first wish. Overall, in the first admission stage, 60\% obtained their first wish, 14\% their second, and 8\% their third. On average, applicants put 6.6 wishes in their preferences list. That year, random choices were used to meet capacity constraints on the first wish for 190 studies \cite{APBStats}. \\


While the selection principle used in selective institutions is commonly agreed on (it is based on applicants high school results), studies with limited capacity have  engendered doubts, incomprehension and even polemics. This is because on the one hand such studies can not legally select applicants on the basis of their results or motivation letters but on the other hand, due to the limited capacity constraint, they need to rank-order applicants in order to fulfil this constraint \cite{CodeEduc}. The ministry has responded to specify that priority is given to applicants of the academy and to ensure that applicants preferences are then used in order to have a first ranking. As described in Section \ref{sec:algo:rank}, when this is not enough to meet the capacity constraint, the algorithm uses random choices between equivalent candidates.

Reacting to the pressure of parents, student associations and their lawyers, the ministry has recently released the portion of the source code concerning studies with limited capacity \cite{LeMondeSourceAPB16,SourceAPB2016}. In the light of this code it appears that all is not described in the algorithm given above and officially specified in \cite{APBAlgoCrit}. For example applicants who passed their Baccalaur\'eat abroad are processed first, which gives them absolute priority over all other applicants. This fact was never documented or presented by the ministry before.\\

As we have previously stressed, rank-ordering applicants in studies with limited capacity is troubling for applicants. In the light of comments posted on the net, the use of randomness and the fact that such studies consider applicants preferences to provide their rank-ordering generate frustration and incomprehension: 
\begin{itemize}

\item {\it ``If I am told that my future is based on chance, I would be surprised.''} 

\item {\it ``Should I first put the study that I really like, or the one where I have better chance of entering?''}

\item {\it ``Do I have a chance of getting a place in a selective study, or in this study with limited capacity located in another academy than that where I pass my bac?''}

\item {\it ``What is the weight of my wish number 1 if it concerns studies with limited capacity?''}

\item {\it ``I do not understand why only the best wishes are offered to the candidates. Why is it not possible to be offered all top-k (e.g. top-3) wishes?''} 

\end{itemize}

\section{\mytitle}

In APB applicants must provide a rank-order (preferences list) over (maximally) 24 studies. This rank-order is a strict total order. This has the unfortunate consequence to force applicants distinguish between all rank-ordered studies which might not be faithful to real applicants' preferences.

In the following we show how a forced linearizing of preferences and random decisions can lead to sub optimal allocation of applicants to institutions. At contrary, we show how the introduction of partial preorders:

\begin{enumerate}
\item gives applicants more expressiveness in their wishes,  
\item allow applicants to delay their definite preferences, and 
\item reduce exposure to random decisions in matching algorithms.   
\end{enumerate}

\subsection{The Benefits of Partial Ordering}
From the analysis of the APB source code for non selective limited capacity studies, it appears that the algorithm is successively executed for each study to produce a ranked list of candidates. Those ranked lists are then used to define institutions/studies' preferences in the Gale and Shapley algorithm.

\begin{example}\label{ex2}
Suppose we have two studies, $S_1$ and $S_2$ with limited capacity equal to 1, and a third study $S_3$ with unlimited capacity. All these studies are located in the same academy $A$. Suppose we have two candidates $C_1$ and $C_2$ in the academy A having the following true preferences lists:

\begin{itemize}
\item[] $C_1: S_1 \succ S_2 \succ S_3$
\item[] $C_2: S_1 \succ S_3$, $S_2 \succ S_3$
\end{itemize}

APB forces the use of strict total orders, and let us assume that as a consequence, $C_2$ decides for $S_1 \succ S_2 \succ S_3$. From now on, the two candidates become indistinguishable for the algorithms. They have the same preferences and live in the same academy. We show below how this leads to sub optimal decisions.\\

Limited capacity studies $S_1$ and $S_2$ have $C_1$ and $C_2$ as candidates. They apply the algorithm  given in Section \ref{sec:algo:rank}, and randomly order candidates. Let us assume that $S_1$ preferences list is $C_2 \succ C_1$, and that $S_2$'s list is $C_1 \succ C_2$.

The Stable Marriage algorithm (see Algorithm \ref{Alg:SM}) will propose its first choice to each institution, resulting in $S_1$ selecting $C_2$ and $S_2$ selecting $C_1$. This means that $C_1$ has her second wish and $C_2$ her first one.\\

Now let us assume that partial orders are introduced in APB. This allows $C_2$ to use her true preferences, $S_1 \succ S_3$, $S_2 \succ S_3$.

\begin{enumerate}

\item
While creating the preferences list for $S_1$, we can now distinguish between $C_1$ and $C_2$, indeed, the second candidate has two top-level wishes while the first candidate has only one. In this case it is reasonable for $S_1$ to give priority to $C_1$.  
\item
While creating preferences for $S_2$, $C_2$ is preferred to $C_1$.
\end{enumerate}

With these new preferences lists, the Stable Marriage procedure can now allocate applicants to institutions: $C_1$ to $S_1$, and $C_2$ to $S_2$. This time the solution gives her first wish to each applicant, and randomness does not impact the selection process.\\

We can now introduce a third candidate $C_3$ from the same academy A. We have:

\begin{itemize}
\item[] $C_1: S_1 \succ S_2 \succ S_3$
\item[] $C_2: S_1 \succ S_3$, $S_2 \succ S_3$
\item[] $C_3: S_2 \succ S_1 \succ S_3$
\end{itemize}

This time the rational is for $S_2$ to give priority to $C_3$, and as a consequence, allocate $C_2$ to $S_3$. Again, this is done without using randomness.

\end{example}

In this example, we have seen that partial orders allowed to more finely distinguish between applicants, and as a consequence, saved the process from random decisions. Partial preorders allow to have ties in applicants preferences which offers more flexibility and allows a better assignment of applicants. 

However, we have also seen that by using a tie in her preferences, $C_2$ is paying the price of being considered after applicants able to express more specific choices. This priority is coherent with AI principles in preferences handling which give priority to more specific preferences \cite{Pe90,Bo94,DuKaPr04a}. Moreover this priority is intuitively meaningful. In fact indifference/incomparability reflects weak preference. Therefore the less an applicant expresses  indifference/incomparability, the more specific her preferences are; thus she deserves to get precedence over less specific applicants.  

Now since the relaxed preferences expressed by an applicant helped the system to minimize the impact of random decisions, we need to give the latter a concrete personal benefit. We propose, when possible, to provide those applicants with the multiple allocation choices in their relaxed constraints. Going back to the previous example, and this time assuming institutions with capacity set to 2. $C_2$, would be proposed to choose between a seat in $S_1$ or in $S_2$. 
This introduces the possibility of a delayed-choice in the APB process. As presented before, initial preferences are expressed in January, and allocations proposed in June-July (see Section \ref{subsec-description-APB}). One can assume that those months could authorize the applicant to refine her thinking and hopefully be able to finally decide between several allocations.

\subsection{Dealing with Partially Ordered Students' Preferences}
We have stressed in the previous section the advantage of offering more flexibility for applicants in expressing their preferences. In particular we do aim at allowing applicants to express their preferences in the form of {\it partial preorders}, namely applicants are allowed to express that some studies are equally desired. The example provided in the previous sub-Section suggests that when a study is considering applicants, knowing that some of these applicants have other equally preferred studies allows to more finely distinguish applicants, and as a consequence minimize the impact of randomness.

More generally, at each step of the algorithm we  rank-order the list of applicants who rank-ordered the study at hand $(r_i,a_j)$, where $r_i$ and $a_j$ respectively stand for `relative rank $i$'' and ``absolute rank $j$'', according to the number of studies that they have rank-ordered $(r_i,a_j)$. The smaller is the number, the better rank-ordered the applicant is. Lastly, randomness is used as an extreme mean to break ties on applicants having the same number of $(r_i,a_j)$-rank-ordered studies. Returning to Example \ref{ex2}, precedence is given to $C_1$ over $C_2$ because the former has only one wish, namely $S_1$ which is rank-ordered $(1,1)$ while the latter has two studies, namely $S_1$ and $S_2$ that are rank-ordered $(1,1)$.\\ 

Technically speaking, given applicant's preferences in the form of a partial preorder we need to know the rank of each study w.r.t. that partial preorder. In Example \ref{ex2}, $S_1$ and $S_2$ needed to know which studies are rank-ordered first by $C_1$ and $C_2$. These studies are $\{S_1\}$ and $\{S_1,S_2\}$ respectively. 

The idea behind this reasoning is to linearize the partial preorder in order to rank-order equivalence classes. Consider again Example \ref{ex2} and  applicant $C_2$. We have that $S_1$ and $S_2$ are rank-ordered first (meant to have an equal preference for $C_2$) and $S_3$ second. This rank-order is a total preorder, $\succeq_t=(E_1,E_2)=(\{S_1,S_2\},\{S_3\})$, which extends the initial partial order provided by $C_2$. \\

\subsubsection{Optimistic vs Pessimistic Reasoning}\label{subsec:opt-pess-reasoning}
We may have multiple total preorders extending a partial preorder. The better is to deal with only one such a total preorder in order to guarantee a robust treatment of applicants' preferences \cite{Bo94}. For this purpose we will use insights from non-monotonic reasoning \cite{Ya83} which are referred to as optimistic reasoning and pessimistic reasoning \cite{Ka11}. More precisely we distinguish between the following two ways of constructing a total preorder which extends a partial preorder. Let $\succeq_t=(E_1,\cdots,E_n)$ the total preorder extending a partial preorder $\succeq_p$ over a set of objects $\mathcal{O}$.
\begin{enumerate}

\item {\bf Optimistic reasoning:}\\
Put in $E_1$ objects in $\mathcal{O}$ which are not strictly less preferred to any other object in $\mathcal{O}$ w.r.t. $\succeq_p$. Remove elements of $E_1$ from $\mathcal{O}$. Repeat this process until $\mathcal{O}$ is empty. Then, $\succeq_t=(E_1,\cdots,E_n)$. 
  
\item {\bf Pessimistic reasoning:}\\
Put in $E'_1$ objects in $\mathcal{O}$ which are not strictly preferred to any other object in $\mathcal{O}$ w.r.t. $\succeq_p$. Remove elements of $E'_1$ from $\mathcal{O}$. Repeat this process until $\mathcal{O}$ is empty. Then, $\succeq_t=(E_1,\cdots,E_n)=(E'_n,\cdots,E'_1)$.

\end{enumerate}
\begin{example}\label{ex3}
Let $\mathcal{O}=\{a,b,c,d,e\}$ be a set of objects. Let $\succeq_p:$ $a\succ b$, $c\succ d\succ e$.\\
Following optimistic reasoning we have $\succeq_t=(E_1,E_2,E_3)=(\{a,c\},\{b,d\},\{e\})$. \\
Following pessimistic reasoning we have $\succeq_t=(E_1,E_2,E_3)=(\{c\},\{a,d\},\{b,e\})$. 
\end{example}
It has been shown that the total preorder $\succeq_t$ is unique following both reasoning lines \cite{Pe90,Bo94,DuKaPr04a}. \\

Optimistic reasoning is gravitation toward the best, namely each object in $\mathcal{O}$ is put in the highest possible rank in $\succeq_t$. On the other hand, pessimistic reasoning is gravitation toward the worst, namely each object is put in the lowest possible rank in $\succeq_t$. Roughly, objects are rank-ordered following optimistic reasoning w.r.t. the number of objects to which they are less preferred. In our example $a$ and $c$ are each less preferred to none  object, $b$ and $d$ are each less preferred to one object while $e$ is less preferred to 2 objects. Following pessimistic reasoning objects are rank-ordered w.r.t. the number of objects which they are preferred to. In our example $c$ is preferred to 2 objects, $a$ and $d$ are each preferred to one object, and $b$ and $e$ are each preferred to none object. 

In AI preference reasoning  optimistic reasoning is applied to negative preferences (called also rejections or constraints) \cite{BeDuKaPr02c}. The principle underlying such preferences is ``what is not excluded is desired''. Pessimistic reasoning is applied to positive preferences which obey the principle ``desired only what is explicitly desired''  \cite{DuHaPr00,BeDuKaPr02c}. 

\subsubsection{Indifference vs Incomparability}
Now that we have stated the benefit of allowing flexible expressions of preferences in terms of partial preorders. It seems important to stress the distinction between indifference and incomparability. In fact the distinction between the two notions is not always understood. It is worth noticing that indifference is an ``information'' that expresses equal preference between objects. On the other hand, incomparability is ``lack of information''. Therefore expressing indifference is more constraining than expressing incomparability. The two notions generally induce different results. 
\begin{example}(Example \ref{ex3} continued)\\
In the previous example $a$ and $c$ were incomparable. Suppose now that they are equally preferred w.r.t. $\succeq_p$. Therefore we have 
$\succeq_p:$ $a\approx c\succ b$, $c\succ d\succ e$. Following optimistic reasoning we have $\succeq_t=(E_1,E_2,E_3)=(\{a,c\},\{b,d\},\{e\})$. Following pessimistic reasoning we have $\succeq_t=(E_1,E_2,E_3)=(\{a,c\},\{d\},\{b,e\})$. 
\end{example}
\subsubsection{Optimistic or Pessimistic Reasoning for APB?}


In an allocation system we aim at satisfying applicants' preferences as best as possible. Therefore we first try to allocate an applicant to her top preferred studies w.r.t. her preferences list. If this is not possible then we look at immediately less preferred studies, and so on. In Example \ref{ex3} if $a$, $b$, $c$, $d$ and $e$ were studies then the allocation system should first try to allocate $a$ or $c$ to the applicant. If none is possible then $b$ and $d$ should be considered. This reasoning line is nothing but optimistic reasoning. Therefore in the rest of this presentation we consider optimistic reasoning to deal with partially ordered applicants' preferences. Algorithm \ref{algo-opt} provides a formal computation of $\succeq_t$ sketched in item 1 above (sub-Section \ref{subsec:opt-pess-reasoning}).

\begin{algorithm}[htbp]
\KwData{A set of objects $\mathcal{O}$, a partial preorder $\succeq_p$ over $\mathcal{O}$.}
\KwResult{A total preorder $\succeq_t=(E_1,\cdots,E_n)$ over $\mathcal{O}$.}
\Begin{
  $\mathcal{S}=\{(o,o'):o,o'\in\mathcal{O},o\succ_p o'\}$  \\
  $\mathcal{D}=\{(o,o'):o,o'\in\mathcal{O}, o\neq o',o\approx_p o'\}$  \\
  $l= 0$ \\
  \While {$\mathcal{O}\neq\emptyset$}
  {
  	$l= l+1$ \\
  	$E_l=\{o_i|o_i\in\mathcal{O},\nexists o_j\in\mathcal{O}\mbox{ s.t. }(o_j,o_i)\in\mathcal{S}\}$  
  
  \For{each $o_i\in E_l$}
   {
   	\lIf{$\exists(o_i,o_j)\mbox{ or }(o_j,o_i)\in\mathcal{D}$ and $o_j\not\in E_l$}{remove $o_i$ from $E_l$}
   }
   \lIf {$E_{l}=\emptyset$} {Stop (inconsistent preference statements)}
   	from $\mathcal{S}$ remove $(o,o')$ if $o\in E_l$  \\
	from $\mathcal{D}$ remove $(o,o')$ if $o,o'\in E_l$  \\
	from $\mathcal{O}$ remove elements of $E_{l}$
   }
\KwRet{$(E_{1},\cdots,E_{l})$}
}
\label{algo-opt}
\caption{Computing a total preorder extending a partial preorder following optimistic reasoning}
\end{algorithm}

One may wonder whether expressing a partial preorder would not give advantage to applicants. For example let us just suppose that an applicant rank-orders all studies first, assuming that she will have a better chance to get her preferred study. This will possibly be the case if the applicant has revealed her true preferences by equally rank-ordering these studies. If this is not the case then she may be allocated to a study that she does not really prefer. This is because, at a given stage, priority will be given to  applicants who have less choices, i.e. applicants who expressed more specific preferences (i.e. expressed less equally preferred studies). Back to Example \ref{ex2} $C_1$ has precedence over $C_2$ because she has only one preferred study, namely $S_1$, while $C_2$ has two preferred studies, namely $S_1$ and $S_2$. An optimal allocation system would allocate $S_1$ to $C_1$ and $S_2$ to $C_2$. Therefore both $C_1$ and $C_2$ get their preferred choice. Now if $C_2$ has not expressed her true preferences and, in reality, prefers $S_1$, she will be disadvantaged by expressing more equally preferred studies than necessary. 

\subsection{Revised APB Algorithms}
We have seen that allowing preorders in applicants preferences helps to distinguish between them. For non selective limited capacity institutions this is crucial since APB relies on random decisions to separate equivalent applicants (see sub-Section \ref{sec:algo:rank}). Selective institutions are not impacted by random decisions. Nevertheless they would definitely benefit from the flexibility offered by partial preorders to express their preferences.

A total order is required by APB's stable marriage procedure\footnote{Remark that in many practical applications, a person may not wish (or be able) to choose between alternatives. As a consequence, the Stable Marriage procedure presented in Section \ref{Alg:SM} has been extended to deal with these cases \cite{DBLP:journals/dam/Irving94}.} which must be able to  gradually exploit preferences lists to converge to a stable solution. Since in general, applicants or institutions can produce disconnected partial orders as their preferences lists, we have presented a general method to compute a total preorder from partial preorders. Considering the intent of preferences expression in APB, we came to the conclusion that optimistic reasoning should be preferred to pessimistic reasoning while computing total preorders.

In the following we show how the use of total preorders as preferences lists impacts the APB algorithm used to automatically create institutions/studies preferences. \\


\begin{algorithm}[htbp]
\KwData{\emph{Institutions}, list of institutions}
\KwResult{For each study $s$ belonging to some institution $i$, \emph{s.Selected} preferences list of applicants}
\Begin{   
  \For {i $\in$ Institutions}{  
    \For {s $\in$ i.Studies}{ 
    		\For(\tcp*[h]{relative ordering}){r $\in$ 1 .. 24}{  
    			\For(\tcp*[h]{absolute ordering}) {a $\in$ 1 .. 24}{ 
    				\For(\tcp*[h]{disjunction size}) {d $\in$ 1 .. maxDisj}{   
					locals = $\{ app \in s.Applicants$ s.t. \\
						$app.Academy = i.Academy$
						and\\
						$app.relative(s) = r$
						and
						$app.absolute(s) = a$ 
						and					
						$app.disj(s) = d$	  \tcp*[h]{s referenced in disjunction of size d}
						$\}$\\
					$randList$ = randomSort(locals)\\
					$add(s.Selected, randList)$ 
				}
			}
  		}
  		
   		\For(\tcp*[h]{relative ordering}){r $\in$ 1 .. 24}{  
    			\For(\tcp*[h]{absolute ordering}) {a $\in$ 1 .. 24}{
    				\For(\tcp*[h]{disjunction size}) {d $\in$ 1 .. maxDisj}{   
					nonLocals = $\{ app \in s.Applicants$ s.t. \\
						$app.Academy \neq i.Academy$
						and\\
						$app.relative(s) = r$
						and
						$app.absolute(s) = a$ 
						and					
						$app.disj(s) = d$	  \tcp*[h]{s referenced in disjunction of size d}
						$\}$\\
					$randList$ = randomSort(nonLocals)\\
					$add(s.Selected, randList)$ 
				}
			}
  		}
  }
  }
\label{Alg:CreatePrefTwo}
\caption{Create Institutions Preferences Using Indifference}
}
\end{algorithm}

We need to generalize to total preorders the APB method which   creates  institutions/studies preferences list according to applicants preferences. 
We want to exploit the knowledge that at a given rank, if applicants $a$ and $b$ prefer a study $t$ but $b$ has also indicated, at the same level, indifference with a study $t'$, $t$ will give priority to $a$. 

Algorithm \ref{Alg:CreatePrefTwo} generalizes the previous idea to increasing degrees of indifference. It outputs for each study a total order on applicants. The constant \emph{maxDisj} reflects some limit on the size of a disjunction. Applicants are considered starting with the ones who gave more specific preferences (no or small disjunction) first.
%

We focused in the previous sections on partial preorders which are the most general form of preferences. Whatever students preferences list being partial or total preorders, when considering these preferences to construct institutions/studies preferences list over students the problem turns out to linearize students preferences lists in the case they are in the form of partial preorders. 
\section{Related Work}

The Stable Marriage problem is based on a very general notion of stability which corresponds to allocations where no individuals can discern any gain from further trade. It became central in cooperative game theory and found applications in a variety of important real-life scenarios \cite{DBLP:journals/jet/RothSU05,NBERw6963}. As a recognition of this large impact, the 2012 Nobel Prize in Economics was awarded to Lloyd S. Shapley and Alvin E. Roth \emph{for the theory of stable allocations and the practice of market design \cite{Nobel}}.


In the domain of education, a large number of U.S. metropolitan areas use some variant of the Gale-Shapley algorithm for high school admission. It all started in 2004, when the New York department of education introduced an applicant-proposing version of the Gale and Shapley algorithm for high school admission. In these systems, total orders are used to express preferences \cite{Abdulkadiroglu05thenew}. Remark that they do not use randomness since institutions are legally entailed to perform selection.

In Artificial Intelligence, several researchers considered the stable marriage problem. In \cite{DBLP:journals/algorithms/GelainPRVW13}, the authors introduce a local search approach which exploits properties of the problems to reduce the size of a state neighborhood. In \cite{DBLP:conf/atal/GelainPRVW10}, male optimality and uniqueness of stable marriages for partially ordered preferences are studied.

In the domain of Constraint Programming, \cite{McCreeshPT16} explores stability in different problems variations including cases where men/women can rank members of their own genders, etc. A constraint encoding of the SM problem with ties and incomplete lists is presented and empirically evaluated in \cite{DBLP:conf/ecai/GentP02}.

In the related field of Satisfiability, parallel solvers can solve, at scale, formulae encoding these NP-complete problems, \cite{DBLP:books/sp/HS2018}, \cite{hamadi:hal-00872792}, \cite{manysat-solver-description}.


\section{Summary and Conclusion}
The Admission Post-Baccalaur\'eat Platform (APB) is used every year to allocate hundreds of thousands first year students to more than twelve thousands university degrees. This is done automatically by matching applicants preferences to university seats. In practice, APB can be unable to distinguish between applicants through their expressed preferences, and as a consequence it can introduce randomness in the matching process. This has created frustration and incomprehension in the public since randomness, even applied as a last mean, is hard to accept in a system used to decide about someone's career. In this work, we have provided a first solution to this problem. We have shown how partial preorders can be introduced to improve applicants expressiveness and consequently reduce exposure to random ordering. Moreover allowing partial preorders permit applicants to postpone their definite preferences. In fact it may be the case that an applicant hesitates between two or more studies in January -- when asked to express her preferences -- but that her preferences become more precise in June-July -- when she has to state her final choice. 


Since in general, applicants or institutions can produce disconnected partial preorders as their preference lists, we have presented a general method to compute a total preorder from partial preorders. Considering the intent of preferences expression in APB, we came to the conclusion that optimistic preferences reasoning should be favored to pessimistic reasoning while computing total preorders.

Our solution has been thoroughly described using examples, theoretical analysis, and new preferences processing algorithms. This was done to allow a quick reuse in some future version of the APB platform. Remarkably, our work solves both minister and applicants concerns. The former has officially acknowledged as an objective the reduction of random choices \cite{APBStats}, and the latter have often raised their voices in public forums to criticize APB's restricted expressiveness and systematic use of randomness.

Although we focused on the APB application our proposal to use partial preorders instead of total orders may be of interest in any preferences-based matching procedures. 

The initial ingredient in the APB platform is applicants' preferences which are total orders over (maximally) 24 studies. In this paper we extended applicants' preferences to deal with partial preorders thus permitting more flexibility. Nevertheless it is well established that humans are generally cognitively unable to provide such (pre)orders because of the cognitive effort required \cite{Slovic95}. Moreover humans generally do not have prior preferences but construct them on the fly \cite{Slovic95}. AI researchers have addressed this problem. In particular instead of asking humans to provide an explicit (pre)order over a set of objects, they consider partial and compact humans preferences. Completion principles are then used to compute the associated pre(order) \cite{Boetal04,Wi04,BiLaWi10,Ka11}. In a future work we intend to extend the APB platform with elicitation of applicants compact preferences as initial input.      

In our societies, more and more decisions are going to be based on the output of automated decision processes. APB is a striking example of these systems whose lack of transparency has been heavily criticized. This has been exacerbated since it touches one of the last symbol of the French republican egalitarian principle: the universal access to university for bacheliers. As a consequence, the pressure put by parents, students, associations and their lawyers has forced the minister to gradually open the black-box and give information on the underlying APB algorithms. As AI researchers, we felt obliged to consider this problem and to provide new algorithms which would improve the APB platform. We hope that more AI researchers will consider the analysis of similar systems whose decisions have direct societal impact.

\bibliographystyle{elsarticle-num}
\bibliography{APB-biblio} 

\end{document}